# FROM LANGUAGE TO VISION: A CASE STUDY OF TEXT ANIMATION

Ping Chen, Richard Alo, Justin Rundell
Department of Computer and Mathematical Sciences
University of Houston – Downtown
One Main Street
Houston, Texas 77002
E-mail: chenp, alor, rundellj1@uhd.edu

*Abstract:* Information can be expressed in multiple formats including natural language, images, and motions. Human intelligence usually faces little difficulty to convert from one format to another format, which often shows a true understanding of encoded information. Moreover, such conversions have broad application in many real-world applications. In this paper, we present a text visualization system that can visualize free text with animations. Our system is illustrated by visualizing example sentences of elementary Physics laws.

## 1. Introduction

Today's culture has been exposed to a rapid growth in multimedia and unparalleled advances in the graphic and visual arts. Visual stimulation bombards the youth of today with images previously only dreamt of. And whether it's tweeting on Twitter or chatting with friends on Facebook, exposure to a vast array of visual stimulation is unavoidable. Consequently, teens have grown accustomed to such visual effects in every aspect of their life. As ever increasing stimuli vie for the attention of today's youth, it becomes all the more apparent that everybody could benefit from similar visual advances. For example, could not a student benefit from a visual example of, for instance, a Physics law, rather than simply reading an example from a book? While applications of free text visualization are endless, there remain numerous challenges that are actively investigated in natural language processing, semantics, and cognitive science [5]. Among these efforts, Genesis Project from MIT proposes the following points of view [1] [6]:

o "Language and vision and motor systems have to talk to each other"

o "Abstract thinking grounds in the physical world"

With these principles in mind, in this paper we focus on visualizing examples of elementary Physics laws since these laws can describe a broad range of real-world entity movement and other critical properties.

## 2. Text Visualization System

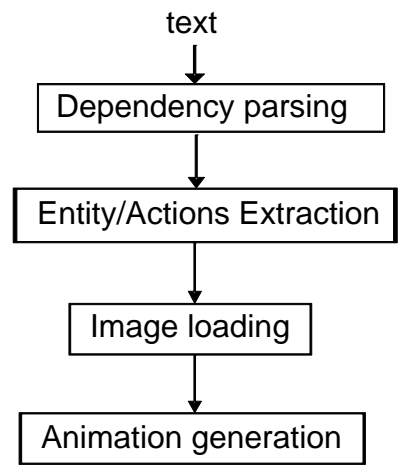

Figure 1 Text Visualization System Configuration

Our text visualization system is shown in Figure 1. Text is parsed first with a dependent parser. Entities and actions, such as person name, location, type of movement, are extracted. Corresponding images are loaded from an image base. Finally these images are integrated into an animation. Generalizing such a process for broad-coverage text visualization brings numerous challenges, which will be analyzed as follows.

### 2.1 Named Entity Recognition

Named entity detection and extraction techniques try to locate and extract the entity names (such as of company, people, locations [7]), monetary amounts, and other similar entities in unstructured text. In early systems usually a domain-specific dictionary and a pattern/rule base are built manually and tuned for a particular corpus. Extraction quality depends on the quality of these external dictionaries and bases, sufficiency of training and consistency of documents within the corpus. Recently more systems utilize context information to deal better with inconsistency among documents, which results in a more robust system. In [4] a semi-Markov model is proposed to make better use of external dictionaries. In [9] a maximum entropy Markov model is introduced to segment FAQ's. Maximum entropy (ME) is also used in [2] to combine diverse knowledge sources. Both hidden Markov model (HMM) and ME can generate statistical models of words and simple word features. In [3] document (not the whole corpus) specific rules are learned for named entities extraction to keep more knowledge of original documents.

Named entity detection focuses on extracting simple terms, hopefully to get some clues for development of general NLP techniques. However, a named entity often has semantic relations with other parts of text, and focusing on only named entities ignores these semantic connections. Our approach is to adopt dependency parsing to accurately extract these relations.

### 2.2 Dependency Parsing

Dependency relations generated by dependency parsing of text can reveal semantic relations among entities [8]. A dependency relation is an asymmetric binary relation between two words, one called head or governor, and the other called dependent or modifier [10]. In dependency grammars a sentence is represented as a set of dependency relations, which normally form a tree that connects all the words in the sentence. For example, "blue sky" contains one dependency relation: "sky -> blue", where "sky" is the head, and "blue" is the dependent. Such a relation will constrain the image selection step as which sky image should be selected. Here is the parsing tree generated from "A rocket ship accelerates in space."

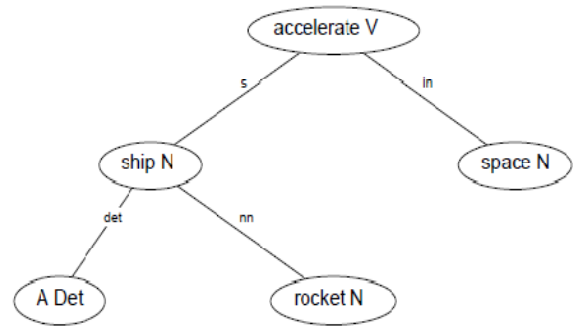

Figure 2 Parsing tree for "A rocket ship accelerates in space."

## 3. Visualizing Physics Laws Animations

Below we traverse through each example in detail, depicting our approach to each animation as well as anything we would change about the animation, if applicable.

### 3.1 Newton's First Law of Motion – Car Crash Example

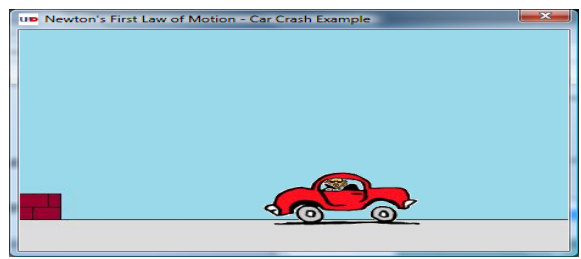

Figure 3 Visualizing "A car crashes into a wall and the driver is ejected."

The animation of a car crash is one of the more complicated animations we constructed. The background of the form is a picture of a street against a blue sky with a brick wall at the leftmost side of the picture. Although it appears that the car is travelling with a man inside, this is not the case programmatically. The man is actually another image moving at the same rate as the car. By drawing the man on the form first and the car second, it appears that the car is moving with the man inside. Once the car collides with the brick wall, the car image is exchanged with an image of a wrecked car and the image of the man inside of the car moves away from the car, appearing to be ejected from the vehicle. We draw the car and the man in three pixel increments by way of a form timer calling a GDI+ paint function.

### 3.2 Newton's Second Law of Motion – Orbiting Moon Example

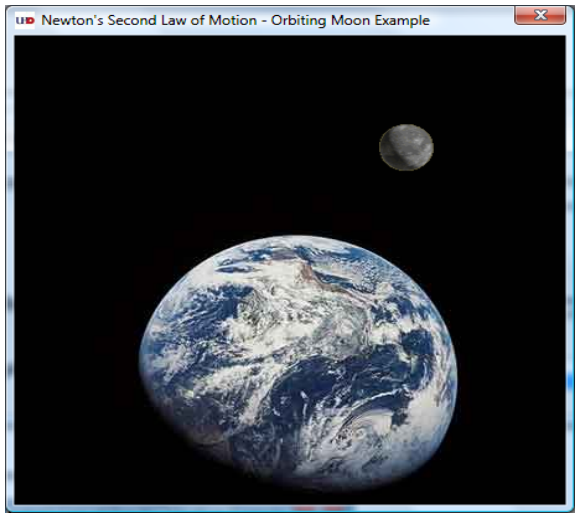

Figure 4 Visualizing "The moon orbits the earth."

The animation of the moon in orbit around the earth proved challenging, as we tried to make the animation as realistic as possible. The background of the form is simply a picture of the earth against a jet black sky and the moon is actually an animated GIF. We devised a way to programmatically deconstruct an animated GIF frame-by-frame in "Newton's Third Law of Motion – Bird Flying Example" and we wanted to employ the technique again in this example. As the program deconstructs the animated GIF into individual frames, the frames are painted to the form once again by a GDI+ paint function called by a form timer. The challenge then becomes how to move the moon in a circle around the earth. We was able to use the mathematical formula of a circle, and by incrementing the position of the moon along that circle by degrees, we could then determine the position of the moon in relation to the earth. Allowing the moon to drift from the viewable area of the form is intentional, adding to the realism of the animation.

### 3.3 Newton's Third Law of Motion – Rocket Example

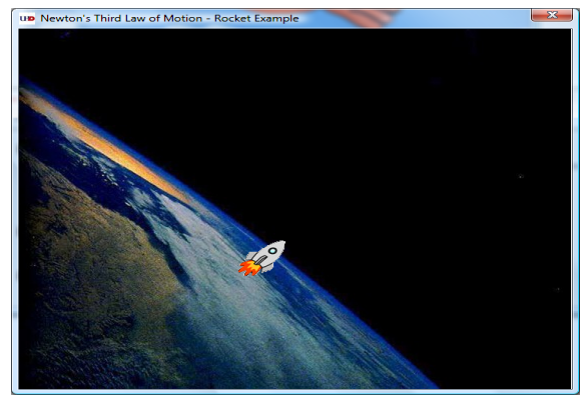

Figure 5 Visualizing "A rocket ship accelerates in space."

The animation of a rocket ship accelerating through space is the first animation where we employed changing the image of the object in motion during the animation. Opening the animation is a stationary picture of a rocket against the backdrop of a planet. After thirty ticks of the form timer, the image changes to a rocket ship at part throttle and the rocket accelerates slightly from its current position. At ninety ticks of the form timer the image of the rocket changes again to that of a rocket ship at full throttle. The rocket now accelerates rapidly off the viewable area of the form. A GDI+ paint function called by the form timer is utilized in this example to draw the image to the form. Had time allowed, we would have liked to have moved the rocket in an arc as opposed to a straight line. We experimented with Bezier Curves to accomplish the arc travel path effect on this example as well as the "Second Law of Thermodynamics – Hurricane Example.

### 3.4 Newton's Law of Gravity – Apple Example

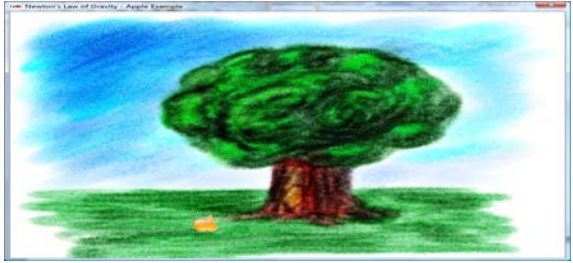

Figure 6 Visualizing "An apple falls from a tree."

The animation of an apple falling from a tree was the first animation we completed for the project. Once we devised a way of displaying the picture of the tree as the form background, we began work on animating the apple. The first hurdle was how to clean up the image of the apple. As with any GIF image, JPEG image or otherwise, an image is always square or rectangular in dimension, regardless of if the image is square or not. In my example, the size of the image was fifty pixels by fifty pixels, making the apple appear as if it were on a white background. Although trivial sounding, we had to devise a method of modifying the image by making the pixels around the apple transparent. Once this was accomplished, my first solution was to animate the apple in a picture box, then move the picture box toward the bottom of the screen. This solution proved to make the image appear to "flash" as the apple fell, which was unacceptable. My second solution was to utilize the GDI+ paint functions to paint the apple at each location as the apple fell from the tree. Additionally, we make the apple appear to accelerate as the apple falls, providing a more realistic animation.

### 3.5 First Law of Thermodynamics – Light Bulb Example

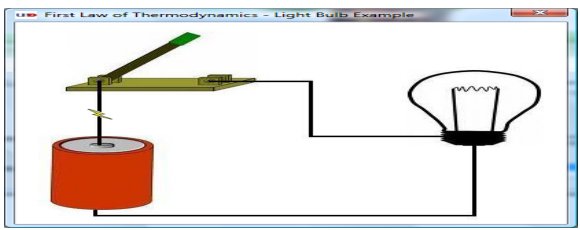

Figure 7 Visualizing "Electricity turns on a light bulb."

For the animation of electricity turning on a light bulb, we wanted to employ a couple of different visual effects. Firstly, we wanted the background image to change to reflect what was happening in the animation at the time. Secondly, we wanted to image of the electric spark to traverse the electrical path animating the objects along the way. The background consists of three different images that update based on where the spark is on the path. The images are of an open circuit, a closed circuit and a closed circuit with the light bulb on. The animation stays synchronized by a form timer and the spark is displayed on the form by a GDI+ paint function.

### 3.6 Second Law of Thermodynamics – Hurricane Example

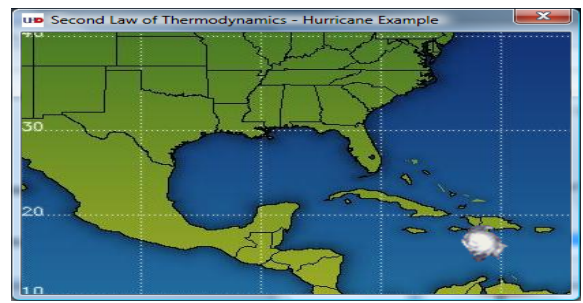

Figure 8 Visualizing "A hurricane forms in the ocean."

For the animation of a hurricane forming in the ocean, we wanted to employ the visual effect of both growth and rotation. The visual effect of growth we learned in the "Boyle's Law – Balloon Example", but for this example, we needed to learn how to rotate an object. The background is an image of the Western Atlantic Ocean and the foreground image is a GIF of a hurricane. As the animation begins, the image of the hurricane is small, increasing in size as it crosses the animation. When the hurricane reaches landfall, the hurricane then begins to shrink, similar to a real-life hurricane. The hurricane grows and shrinks using a GDI+ paint function by specifying a new width and height for the image. A high quality bicubic interpolation mode is set to keep the hurricane as crisp as possible during resizing. The image is rotated using a 2-dimensional drawing function called RotateFlip. This function rotates the image at ninety degree intervals.

### 3.7 Boyle's Law – Balloon Example

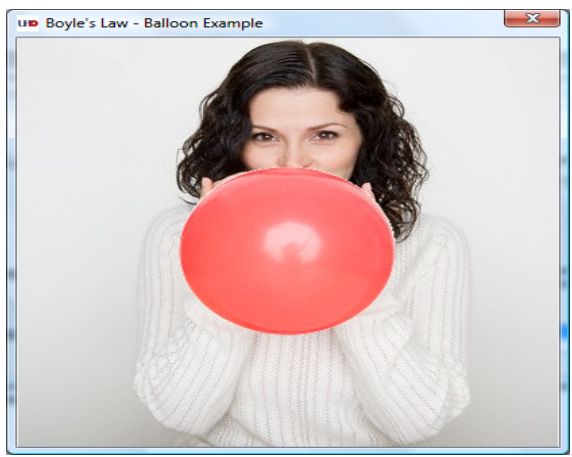

Figure 9 Visualizing "A girl inflates a balloon."

For the animation of a girl inflating a balloon, we wanted to employ growth of the balloon image to obtain the desired visual effect. This required that we learn how to make an image appear to grow. While we could have used a Graphics.FillEllipse function as we did in the "Boyle's Law – Scuba Diver Example", the balloon appears more realistic with the flash of the camera displayed on the balloon surface. The balloon grows by way of specifying new width and height values for the image when the GDI+ paint function is referenced. A high quality bicubic interpolation mode is set to keep the balloon as crisp as possible during resizing. Additionally, we divide the animation into thirds by placing the thread in a sleep state for one second, giving the illusion that the girl is inhaling and then continuing to inflate the balloon.

## 4. Conclusion

Text visualization targets on several core research areas in Artificial Intelligence and has broad application including education, information technology, and many other fields. This paper discusses our early-stage text visualization system. We illustrate our system by visualizing examples of several Physics laws.

## 5. Acknowledgement

This project has been partially supported by DHS 2009-ST-061-C10001 and NSF CNS 0851984.